\pdfoutput=1
\documentclass{article}

\usepackage{amssymb}
\usepackage{graphicx}
\usepackage{multirow}
\usepackage{amsmath}
\newcommand{\biggraphwidth}{0.95\textwidth}

\newcommand{\medgraphwidth}{0.6\textwidth}
\newcommand{\smallgraphwidth}{0.5\textwidth}

\usepackage{algorithm}
\usepackage{algpseudocode}
\usepackage{subfig}
\usepackage{color}
\usepackage{colortbl}

\title{Topic Modeling for Classification of Clinical Reports}

\author
       {Efsun Sarioglu Kayi$^1$, Kabir Yadav$^2$, James M. Chamberlain$^3$,
        Hyeong-Ah Choi $^1$
       \\
       $^1$Department of Computer Science, George Washington University\\
       $^2$Department of Emergency Medicine, Harbor-UCLA Medical Center\\
       $^3$Department of Pediatrics and Emergency Medicine, George Washington University\\
       }

\begin{document}
\maketitle
\begin{abstract}
Electronic health records (EHRs) contain important clinical information about patients. Efficient and effective use of this information could supplement or even replace manual chart review as a means of studying and improving the quality and safety of healthcare delivery. However, some of these clinical data are in the form of free text and require pre-processing before use in automated systems. A common free text data source is radiology reports, typically dictated by radiologists to explain their interpretations. We sought to demonstrate machine learning classification of computed tomography (CT) imaging reports into binary outcomes, i.e. positive and negative for fracture, using regular text classification and classifiers based on topic modeling. Topic modeling provides interpretable themes (topic distributions) in reports, a representation that is more compact than the commonly used bag-of-words representation and can be processed faster than raw text in subsequent automated processes. 
We demonstrate new classifiers based on this topic modeling representation of the reports. Aggregate topic classifier (ATC) and confidence-based topic classifier (CTC) use a single topic that is determined from the training dataset based on different measures to classify the reports on the test dataset. Alternatively, similarity-based topic classifier (STC) measures the similarity between the reports' topic distributions to determine the predicted class. Our proposed topic modeling-based classifier systems are shown to be competitive with existing text classification techniques and provides an efficient and interpretable representation. 
\end{abstract}

\section{Introduction}
Large amounts of clinically important medical data are now stored in electronic health records (EHRs). In addition to simple performance measurements, more advanced uses may include decision support, such as matching prior patient patterns to recommend the need for a certain medical test or therapy. This can improve effectiveness and efficiency by helping the clinician avoid unnecessary or potentially harmful tests  or therapies. However, some of these data are in the form of free text and they need to be processed and coded for better retrieval and analysis by automated or semi-automated systems. 
\\
\\
Topic modeling is an unsupervised technique that can automatically identify themes from a given set of documents and find topic distributions for each of them. Representing reports according to their topic distributions is more compact and therefore, they can be processed faster than raw text in subsequent automated processing. Also, biomedical concepts can be well represented as nouns \cite{Huang:2005fk} and compared to other parts of speech, they tend to specialize better into topics \cite{griffiths2004integrating}. Accordingly, we hypothesized that the topic model representation of patient CT reports consisting of nouns will perform favorably compared to conventional machine learning for automated classification of clinical outcomes.
\\
\\
A preliminary version of this work has been reported in \cite{efsun12d, efsun13a}. In \cite{efsun12d}, the performance of topic vector classification with conventional classifiers was analyzed and in \cite{efsun13a},  aggregate topic classifier (ATC) were introduced using a single dataset. In this study, we introduce two new classifiers, namely, similarity-based and confidence based topic classifiers (STC, CTC), and analyze and compare their performances more thoroughly using two datasets.

\section{Material and Methods}
\label{bgr}
\noindent
 Before going to the results and findings of this research; this section provides the technical background to carry out this research: topic modeling and text classification. 
 For topic modeling, we go over the historical progress in the field by explaining the mainly utilized models and how they differ from each other in Section \ref{bgr_tm}. After that, the two popular classification techniques namely, SVM and decision tree, are explained in Section \ref{bgr_text}. 

\subsection{Topic Modeling}
\label{bgr_tm}
\noindent
Topic modeling is an unsupervised learning algorithm that can automatically discover themes of a document collection. Several techniques can be used for this purpose including Latent Semantic Analysis (LSA) \cite{Deerwester90indexingby}, Probabilistic Latent Semantic Analysis (PLSA) \cite{hofmann1999plsa}, and Latent Dirichlet Allocation (LDA) \cite{blei03}. LSA is a way of representing hidden semantic structure of a term-document matrix in which rows are documents and columns are words/tokens \cite{Deerwester90indexingby} based on Singular Value Decomposition (SVD). One limitation of LSA is that each word is represented as a single point with the same meaning; therefore in this representation, polysemes of words cannot be differentiated. Also, the final output of LSA, which consists of axes in Euclidean space, is not interpretable or descriptive \cite{Hofmann:2001:ULP:599609.599631}. 
\\
\\
PLSA is considered to be a probabilistic version of LSA where an unobserved class variable
is associated with each occurrence of a word in a particular document \cite{hofmann1999plsa}. These classes/topics are then inferred from the input text collection.
PLSA solves the polysemy problem; however it is not considered a fully generative model of documents which can lead to overfitting \cite{blei03}. 
\\
\\
LDA, first defined by Blei et al \cite{blei03}, defines a topic as a distribution over a fixed vocabulary, where each document can exhibit topics with different proportions. 
LDA performs better than PLSA for small datasets because it avoids overfitting and it also supports polysemy \cite{blei03}. Also, in contrast to PLSA, LDA is also considered a fully generative system for documents. Accordingly, LDA is used to generate topic distributions of clinical reports in this study.
\subsection{Text Classification}
\label{bgr_text}
\noindent
Text classification is a supervised machine learning algorithm where each document's category is learned from a pre-labeled set of documents. Decision trees and support vector machines (SVM) are two such classification algorithms. In a decision tree, internal nodes are the selected terms from the vocabulary, the branches are the criteria on the weight of the terms and the leaves represent the classes. SVM, on the other hand, attempts to find a decision boundary between classes that is the farthest from any point in the training dataset. Given labeled training data \((x_t,y_t), t=1,...,N\) where \(x_t \in R^M\) and \(y_t \in \{1,-1\} \), it tries to find a separating hyperplane with the maximum margin \cite{Platt98sequentialminimal}.  
In this study, decision tree and SVM are chosen as classification techniques: Decision tree is preferred due to its explicit rule based output that can be easily evaluated for content validity and SVM performs well in text classification tasks \cite{Joachims:1998:TCS:645326.649721,Yang:1999:RTC:312624.312647}.
\section{Calculation}
\label{experiments}
\noindent
 Our proposed text classification techniques can be used for various domains. However, our main goal for this study was to utilize such techniques for effective classification of clinical reports. As such, radiology reports from various emergency medicine departments were used to evaluate the proposed classifiers performance. The datasets are computed  tomography (CT) imaging reports done for head traumas and they are further explained in the Section \ref{sec:exp_dataset}.  The preprocessing that they go through before any classification is explained in Section \ref{sec:exp_preprocess}  and  the measures that are used to evaluate the performance of these classifiers are explained in Section \ref{sec:exp_evaluation}.  Finally, after explaining the raw text classification of these clinical reports in Section \ref{sec:exp_raw_text},  the proposed topic modeling-based classifiers are explained in Section \ref{sec:exp_tm}.
\subsection{Dataset}
\label{sec:exp_dataset}
\begin{figure*}[tbp]
\centering
\includegraphics[width=\biggraphwidth]{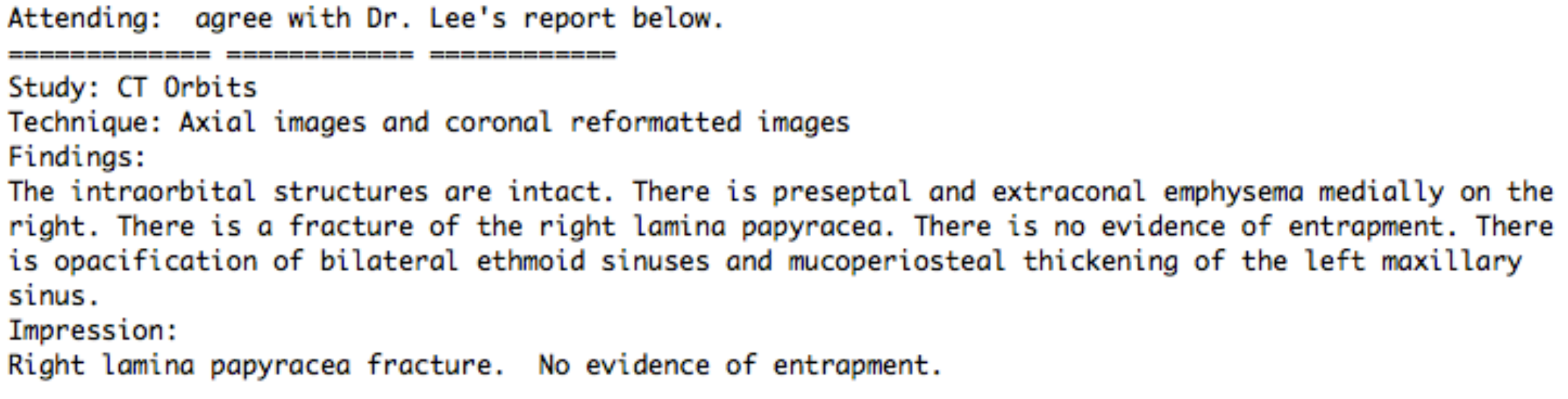}
\caption{ Sample orbital CT report}
\label{fig:report}
\end{figure*}

\begin{figure*}[tbp]
\centering
\includegraphics[width=\biggraphwidth]{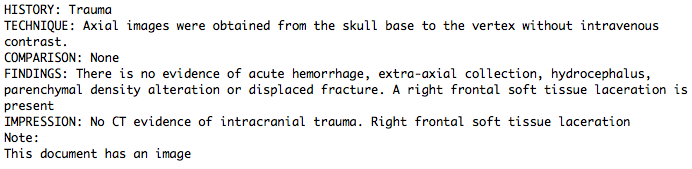}
\caption{ Sample pediatric CT report}
\label{fig:report-ped}
\end{figure*}

\noindent
 This study used prospectively collected patient CT report data previously collected for derivation of a traumatic orbital fracture  clinical risk score \cite{Yadav:2012uq}  and a pediatric traumatic brain injury clinical prediction rule \cite{Kuppermann:2009fk}.  Staff radiologists  dictated  each  CT  report and  the  outcome of interest (either acute  orbital  fracture or findings consistent with traumatic brain injury) was extracted  by a trained  data abstractor. Among the 3,705 orbital CT reports,  3,242 were negative and 463 were positive.  Among the 2,126 pediatric head CT reports, 1,973 were negative and 153 were positive. Figures \ref{fig:report} and \ref{fig:report-ped}  show sample reports from the orbital and pediatric datasets respectively.

\subsection{Evaluation}
\label{sec:exp_evaluation}
\noindent
 In this section, the measures used to evaluate 
 the classification algorithms
 are explained.
\noindent
 Once a classifier is built, its performance is evaluated on a separate dataset. To prevent overfitting, only a subset of the dataset, called the training dataset 
 was used to train the classifier. Its effectiveness was then measured in the remaining unseen documents in the testing set. Also, to effectively measure a classifierÕs success, training and testing datasets with different proportions were prepared: 75\%, 66\%, 50\%, 34\%, and 25\%. 
 These training and test datasets were randomized and stratified to make sure each subset is a good representation of the original dataset in terms of class distribution. The orbital dataset has a positive class ratio of 12,5\% and the pediatric dataset has a positive class ratio of 7,2\%.
To evaluate the classification performance, \emph{precision, recall}, and \emph{F-score}  measures were used. For binary classification, possible cases are summarized in Table \ref{tab:binary_classification}  and Equations \ref{eq:precision} and \ref{eq:recall}  present how \emph{precision} and \emph{recall}  are calculated.
\begin{table}[!htbp]
\centering
\caption{Confusion matrix} 
\vspace{10pt}
\label{tab:binary_classification}
\footnotesize	
\begin{tabular}{cc|c|c|c}
\cline{3-4}
& & \multicolumn{2}{|c|}{\bf Predicted class} \\ \cline{3-4}
& & \bf Positive & \bf Negative \\ \cline{1-4}
\multicolumn{1}{|c|}{\multirow{2}{*}{\bf Actual Class}} &
\multicolumn{1}{|c|}{\bf Positive} & True Positive (TP) & False Negative (FN)  &     \\ \cline{2-4}
\multicolumn{1}{|c|}{}                        &
\multicolumn{1}{|c|}{\bf Negative} & False Positive (FP) & True Negative (TN) &     \\ \cline{1-4}
\end{tabular}
\end{table}
\begin{equation}\label{eq:precision}
    Precision = \frac{TP}{TP+FP}
\end{equation} 
\begin{equation}\label{eq:recall}
    Recall = \frac{TP}{TP+FN}
\end{equation} 
 F-score is calculated as an equally weighted harmonic mean of \emph{precision} and \emph{recall} (See Equation \ref{eq:f-score}):
\begin{equation}\label{eq:f-score}
    \operatorname{F-score} = \frac{2 \times Precision \times Recall} {Precision + Recall}
\end{equation} 
\subsection{Preprocessing}
\label{sec:exp_preprocess}
\noindent
Text data must be converted to a suitable format for automated processing. One common way of doing this is bag-of-words (BoW) representation where each document becomes a vector of its words/tokens. The entries in this matrix could be binary stating the existence or absence of a word in a document or it could be weighted according to the number of times a word exists in a document. For this study, using term weights produced slightly better classification results than other options. Frequent words were also removed from the vocabulary to limit its size. In addition, these frequent words typically do not add much information; most were stop words such as \emph{is, am, are, the, of, at, and}.  Other preprocessing tasks such as stemming was also explored; however, they did not have a significant effect on the classification performance. 
\subsection{Raw Text Classification of Clinical Reports}
\label{sec:exp_raw_text}
\begin{figure}[tbp]
\centering
\includegraphics[width=\biggraphwidth]{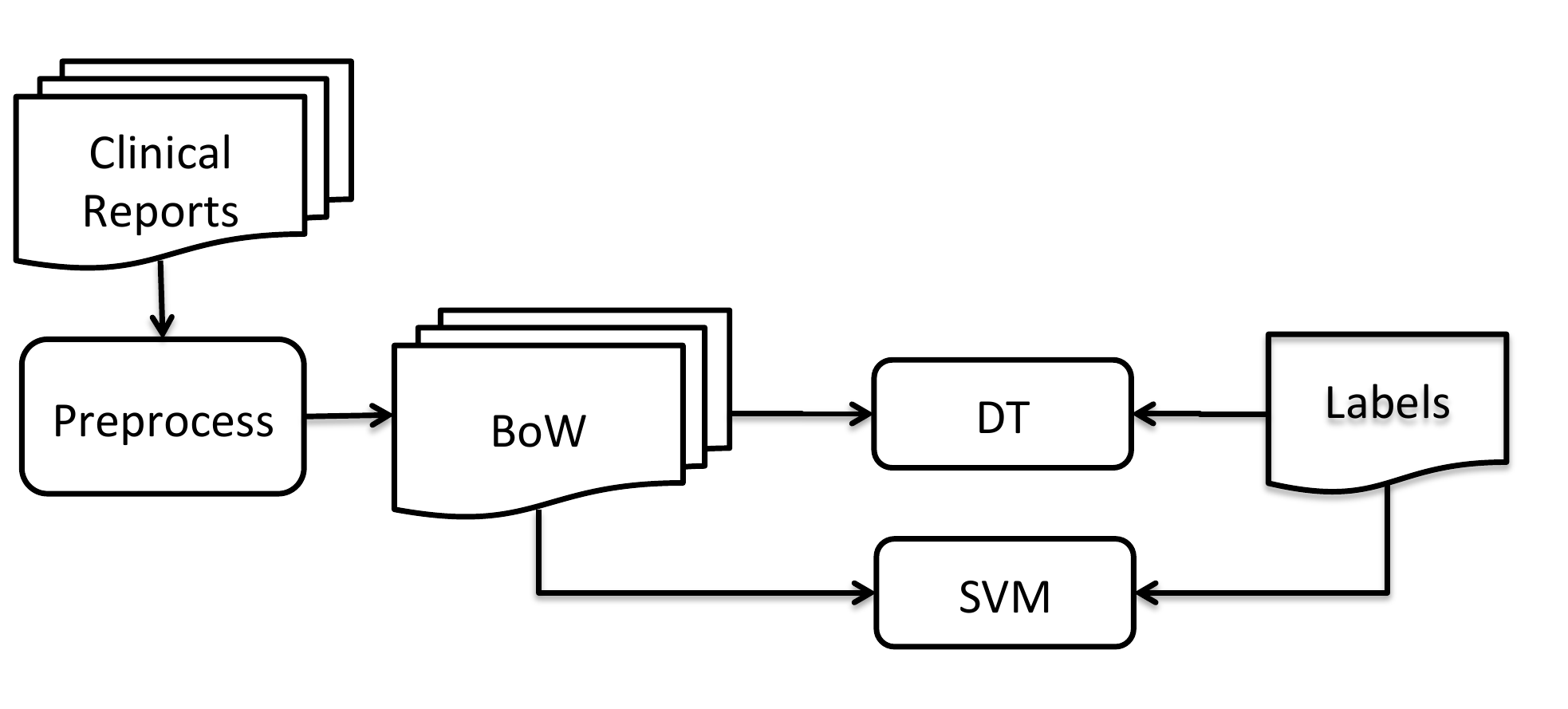}
\caption{System overview for raw text classification}
\label{fig:overviewtext}
\end{figure}

\noindent
Raw text of clinical reports were classified by conventional classification techniques as shown in Figure \ref{fig:overviewtext}. After preprocessing, the raw text files were combined with their associated outcomes and classified using SVM and decision tree in Weka. Weka is a collection of machine learning algorithms for data mining tasks written in Java \cite{hall09}.
\subsection{Topic Modeling-based Classification of Clinical Reports}
\label{sec:exp_tm}
\begin{figure}[tbp]
\centering
\includegraphics[width=\biggraphwidth]{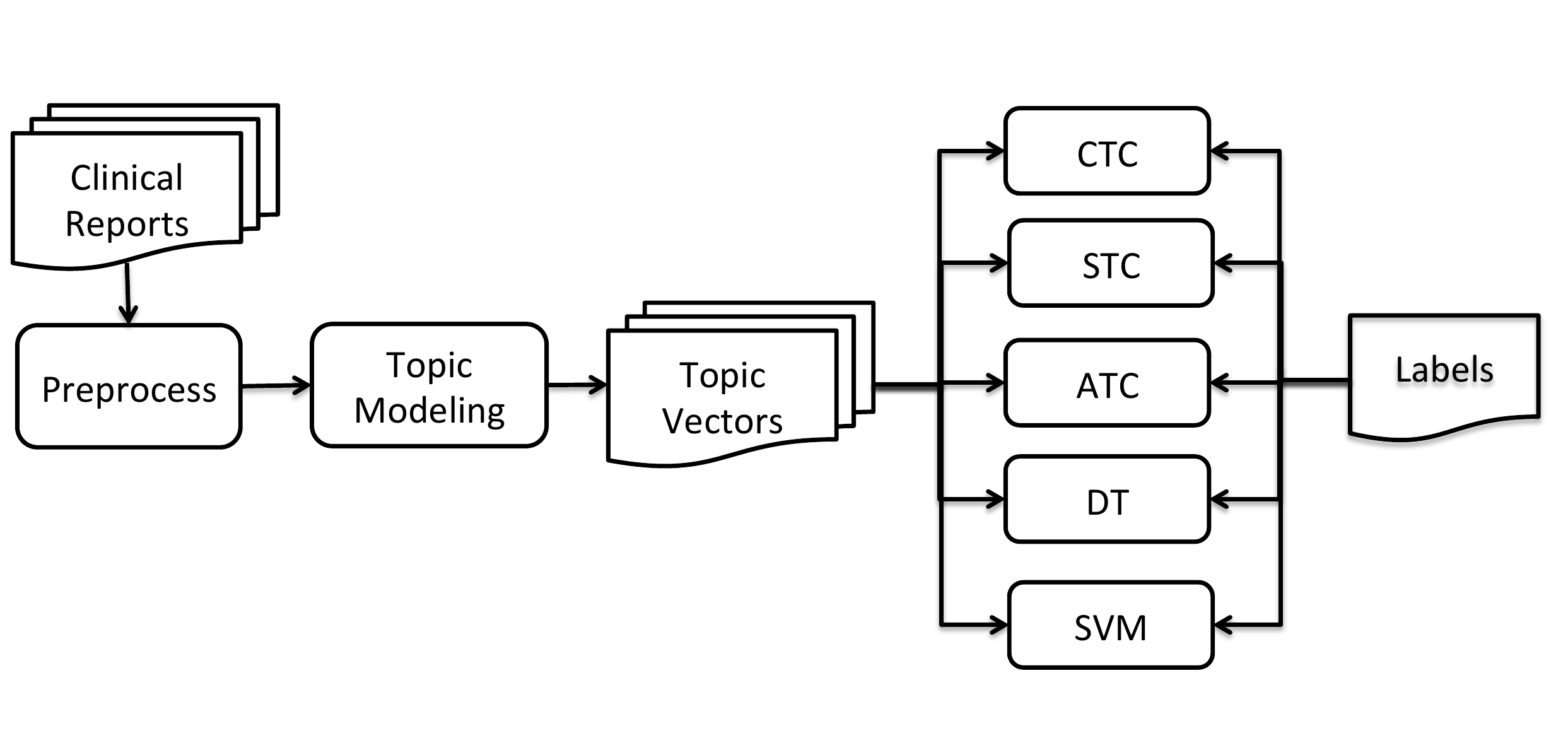} 
\caption{System overview for topic modeling-based classification}
\label{fig:overviewall}
\end{figure}

\noindent
Clinical reports were classified by topic modeling-based classification techniques as shown in Figure \ref{fig:overviewall}. As discussed  Section  \ref{bgr_tm}, we chose LDA to generate the topic models of clinical reports because it is a generative probabilistic system for documents and it is robust to overfitting. The Stanford Topic Modeling Toolbox (TMT) \cite{stanfordTMT}
was used to conduct the experiments. It is an open source software providing ways to train and infer topic models for text data. 
\subsubsection{Topic Vector Classifier}
\label{exp_tv}
\noindent
In topic vector classifier, a topic model of all of the reports were built and the topic distribution of each report was used to represent them in the form of topic vectors. This could be considered as an alternative representation to bag-of-words (BoW), in which terms are replaced with topics and entries for each report  show the probability of a specific topic for that  report. Representation as topic vectors is more compact than BoW because the vocabulary for a text collection usually has thousands of entries, whereas a topic model is typically built with a maximum of hundreds  of topics. These topic vectors were then classified via conventional classification algorithms, e.g., SVM and decision tree (See Algorithm \ref{lst:tvc}). 
\begin{algorithm} 
\begin{algorithmic}
\State learn the topic model for the documents
\State merge the documents in topic vector representation with their classes 
\State train decision tree and SVM using documents represented as topic vectors
\end{algorithmic}
\caption{Topic Vector Classifier}
\label{lst:tvc}
\end{algorithm}
\subsubsection{Confidence-based Topic Classifier (CTC)}
\label{exp_ctc}
\noindent
In this classifier, after the topic model is learned, a single topic is chosen that has the biggest confidence \cite{Agrawal:1993:MAR:170035.170072} for a class. The \emph{confidence} \eqref{eq:confidence} of a topic \emph{X} for a class \emph{Y} is calculated as the \emph{support} \eqref{eq:support} of the topic and the class together divided by the \emph{support} of the topic itself. Using this topic, the predictions for the test dataset are made as shown in  Algorithm \ref{lst:ctc}.
\begin{algorithm} 
\begin{algorithmic}
\State learn the topic model for the documents in the training dataset
\State merge the documents in topic vector representation with their classes 
\State calculate the \(conf(T \Rightarrow C)\) for each topic \emph{T} and class \emph{C} in the training dataset 
\State find the topic \emph{t} with the biggest confidence for the positive class
\State pick a threshold for \emph{th} for the chosen topic \emph{t}
\ForAll{documents in the testing dataset}
	\State infer the document's topic distribution 
	\State find its value \emph{v} for the chosen topic \emph{t}  
	\If {\emph{v}  \textgreater \emph{th}} 
		\State predict as positive 
	\Else
		 \State predict as negative
	\EndIf
\EndFor
\end{algorithmic}
\caption{Confidence-based Topic Classifier (CTC)}
\label{lst:ctc}
\end{algorithm}
\begin{equation}\label{eq:confidence}
conf(X \Rightarrow Y) = \frac{supp(X \cup Y)}{supp(X)}
\end{equation}
\begin{equation}\label{eq:support}
supp(X)= \frac {N_X}{N}
\end{equation}
\subsubsection{Similarity-based Topic Classifier (STC)}
\label{exp_stc}
\noindent
In this classifier, the topic model was learned on the training datasets and the average of topic distributions for each class was calculated. For a document in the testing dataset, its topic distributions were inferred and the class that was the most similar to it was assigned as its predicted class (See Algorithm \ref{lst:ctc}). To calculate the similarity, the cosine measure was used. Given two vectors \emph{x} and \emph{y}, the cosine of the angle between them can be calculated as in Equation \ref{eq:cos}. Its value ranges between 0 and 1 and the more similar the vectors the higher the cosine score is. In this case, one vector represents the average topic distribution for a given class and another vector represents the topic distribution of a test document.
\begin{algorithm} 
\begin{algorithmic}
\State learn the topic model for the documents in the training dataset
\State merge the documents in topic vector representation with their classes 
\State calculate the average topic distribution of each class 
\ForAll{documents in the testing dataset}
         \State infer the document's topic distribution 
	\ForAll{classes}
		\State calculate the similarity between the document's topic distribution and average topic distribution of the class 
	\EndFor
	\State assign the class that is most similar to the document as predicted class
\EndFor
\end{algorithmic}
\caption{Similarity-based Topic Classifier (STC)}
\label{lst:stc}
\end{algorithm}
\begin{equation}\label{eq:cos}
Similarity = cos(\theta) = \frac { x\cdot y}{|x||y|}
\end{equation}
\subsubsection{Aggregate Topic Classifier (ATC)}
\label{exp_atc}
\noindent
With this approach, a representative topic vector for each class was composed by averaging their corresponding topic distributions in the training dataset. A discriminative topic was then chosen so that the difference between positive and negative representative vectors is maximum as shown in Algorithm \ref{lst:atc}. The reports in the test datasets were then classified by analyzing the values of this topic and a threshold was chosen to determine the predicted class. This threshold could be chosen automatically based on class distributions if the dataset is skewed or cross validation methods can be applied to pick a threshold that gives the best classification performance in a validation dataset. This approach is called Aggregate Topic Classifier (ATC) since training labels were utilized in an aggregate fashion using an average function rather than individually. 
\begin{algorithm} 
 \begin{algorithmic}
\State learn the topic model for the documents in the training dataset
\State merge the documents in topic vector representation with their classes 
\State calculate the average of topic distributions of each class 
\State pick the topic \emph{t} whose difference between the average of classes is maximum 
\State pick a threshold \emph{th}  on the selected topic \emph{t} 
\ForAll{documents in the testing dataset}
	 \State infer the document's topic distribution
	 \State find its value \emph{v} for the chosen topic \emph{t}  
	\If { \emph{v} \textgreater \emph{th} }	
		\State predict as positive
	\Else 
		\State predict as negative
\EndIf
\EndFor
\end{algorithmic}
\caption{Aggregate Topic Classifier (ATC)}
\label{lst:atc}
\end{algorithm}
\section{Results}
\label{experiment_results}
\noindent
 The main goal of this study is to analyze and optimize clinical text classification. As a starting point, raw text of clinical reports were classified by well-known conventional classification algorithms. Alternatively, topic modeling of the corpora was used as a compact representation of the clinical reports and classifiers were built using this representation in various ways. The classification results using the proposed topic model-based classifiers are presented according to the evaluation techniques explained in Section \ref{sec:exp_evaluation} .
\subsection{Raw Text Classification Results}
\label{exp_btc}
\begin{figure*}[tp]
    \centering
    \subfloat[Precision
        \label{fig:textPrecisionOrbital}]{
        \includegraphics[width=\smallgraphwidth]{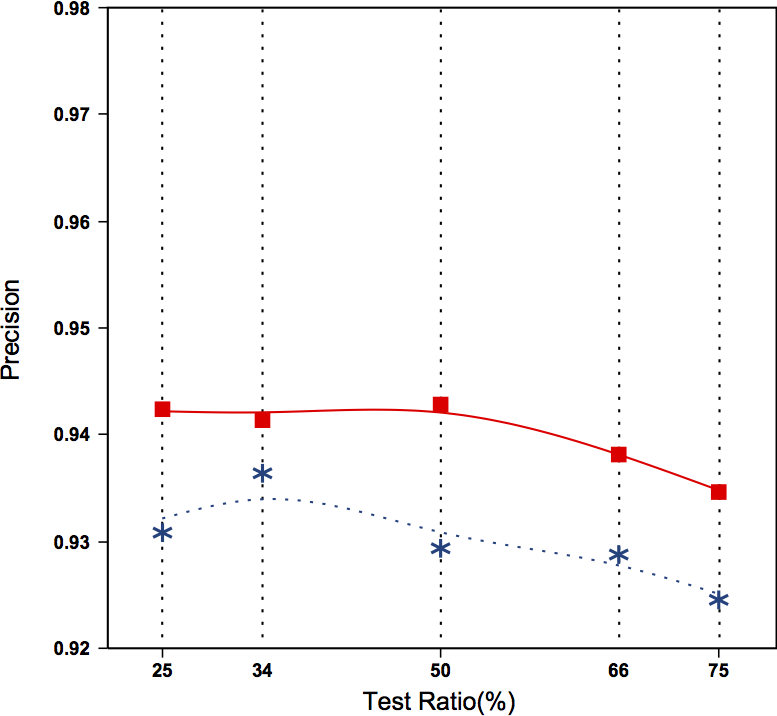}
    }
    \subfloat[Recall
        \label{fig:textRecallOrbital}]{
        \includegraphics[width=\smallgraphwidth]{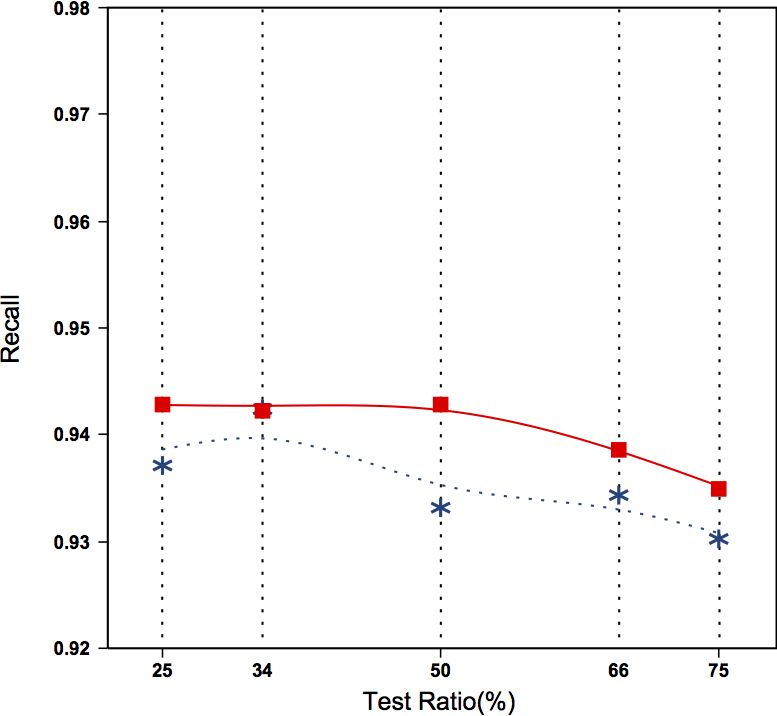}
    }
    \qquad
    \subfloat[F-score
        \label{fig:textFscoreOrbital}]{
        \includegraphics[width=\medgraphwidth]{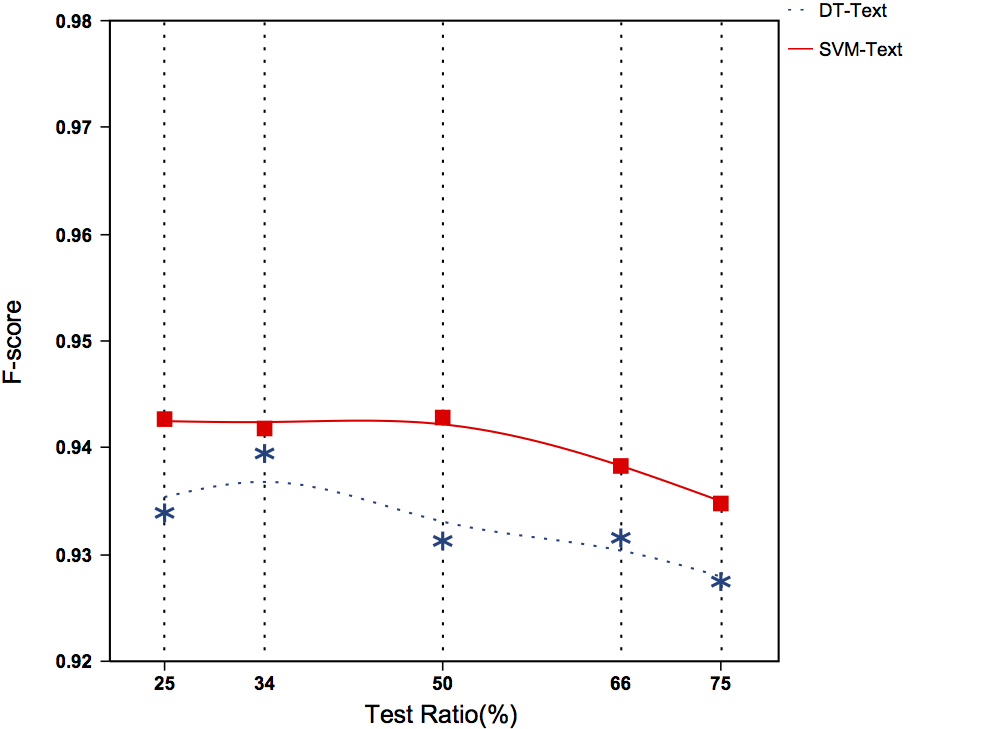}
    }
    \caption{Raw text classification performance for the orbital dataset}
    \label{fig:orbital-text}
\end{figure*}

\begin{figure*}[tp]
    \centering
    \subfloat[Precision
        \label{fig:textPrecisionPediatric}]{
        \includegraphics[width=\smallgraphwidth]{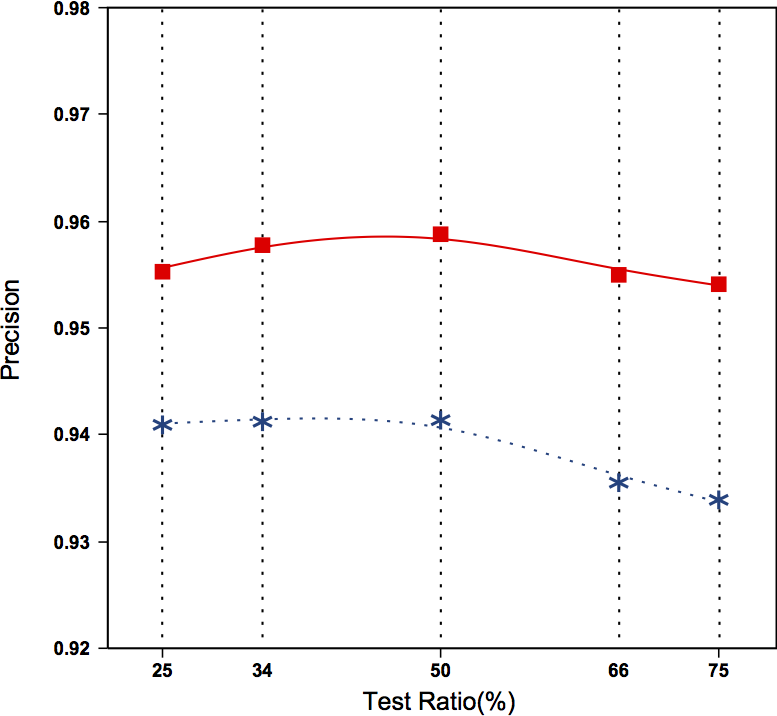}
    }
    \subfloat[Recall
        \label{fig:textRecallPediatric}]{
        \includegraphics[width=\smallgraphwidth]{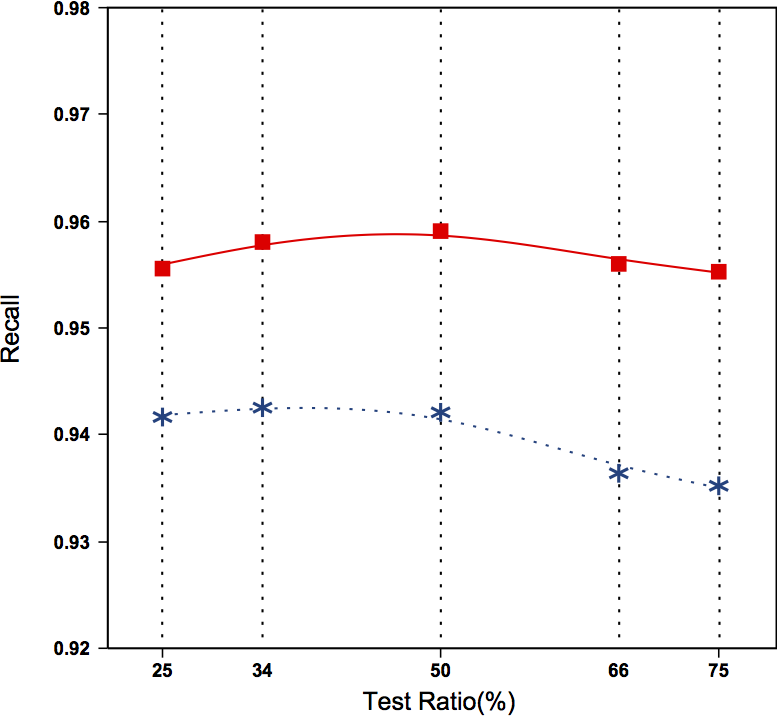}
    }
    \qquad
    \subfloat[F-score
        \label{fig:textFscorePediatric}]{
        \includegraphics[width=\medgraphwidth]{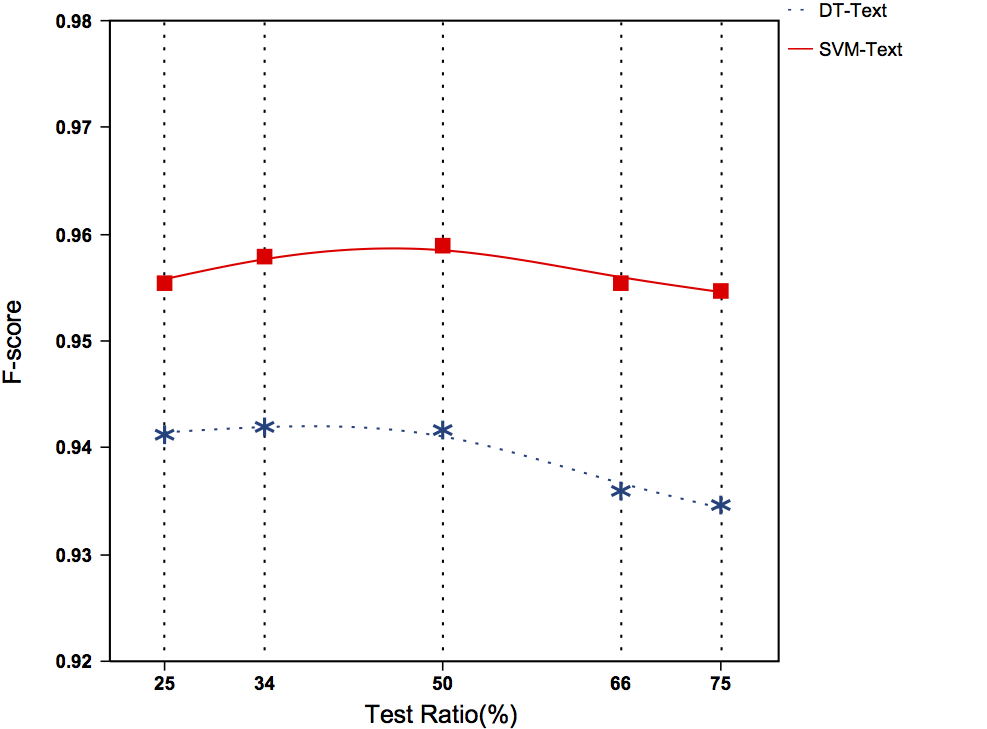}
    }
    \caption{Raw text classification performance for the pediatric dataset}
    \label{fig:pediatric-text}
\end{figure*}

\noindent
Raw text of clinical reports were preprocessed and classified using decision tree (DT) and SVM 
and they are graphically illustrated in Figures \ref{fig:orbital-text} and \ref{fig:pediatric-text} for the orbital and pediatric datasets respectively.  
SVM  performs better than decision tree consistently for different training and testing proportions and for both datasets.
\subsection{Topic Modeling-based Classification Results}
\label{exp_btc}
\begin{figure*}[tp]
    \centering
    \subfloat[Precision
        \label{fig:topicPrecisionOrbital}]{
        \includegraphics[width=\smallgraphwidth]{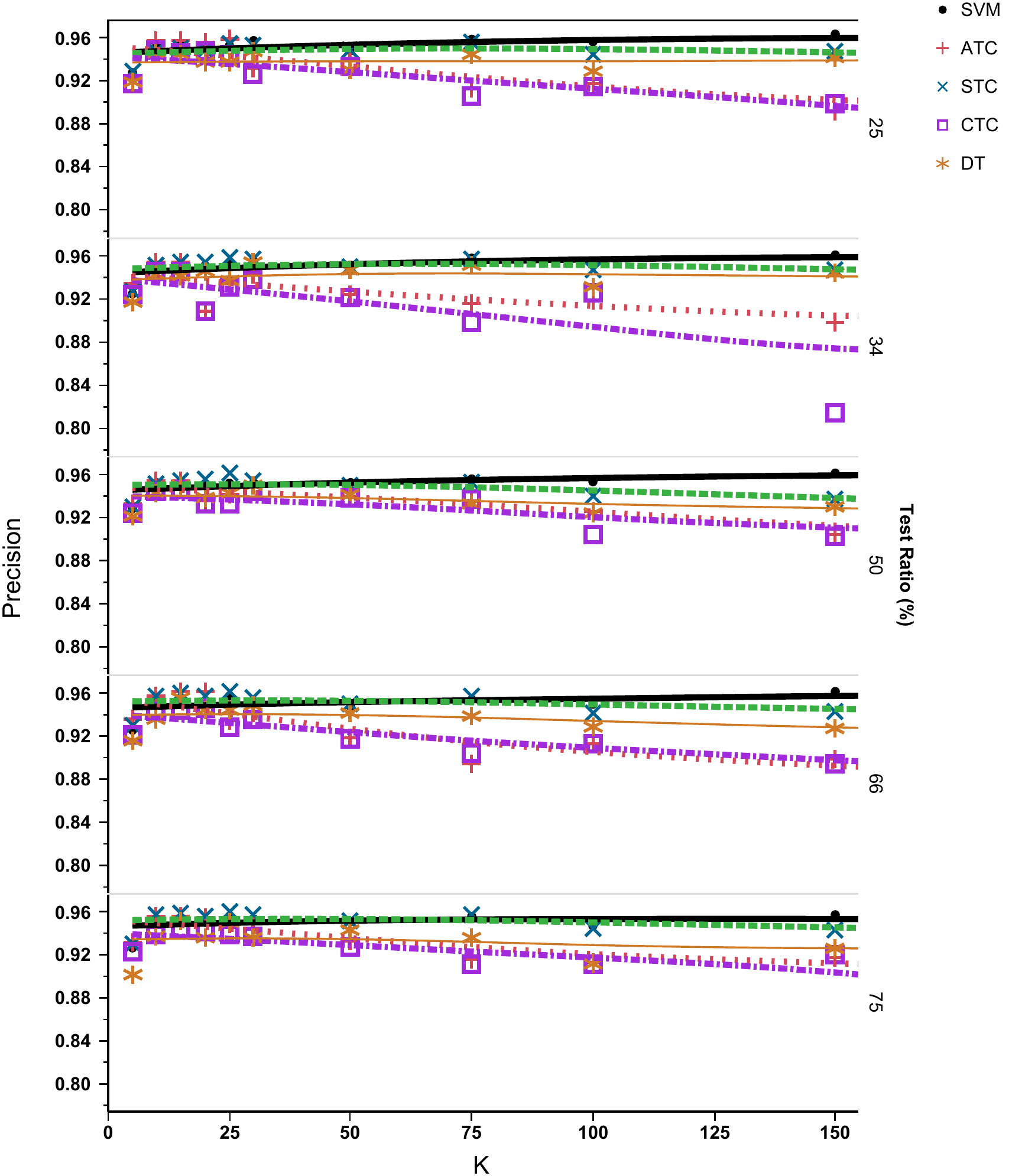}
    }
    \subfloat[Recall
        \label{fig:topicRecallOrbital}]{
        \includegraphics[width=\smallgraphwidth]{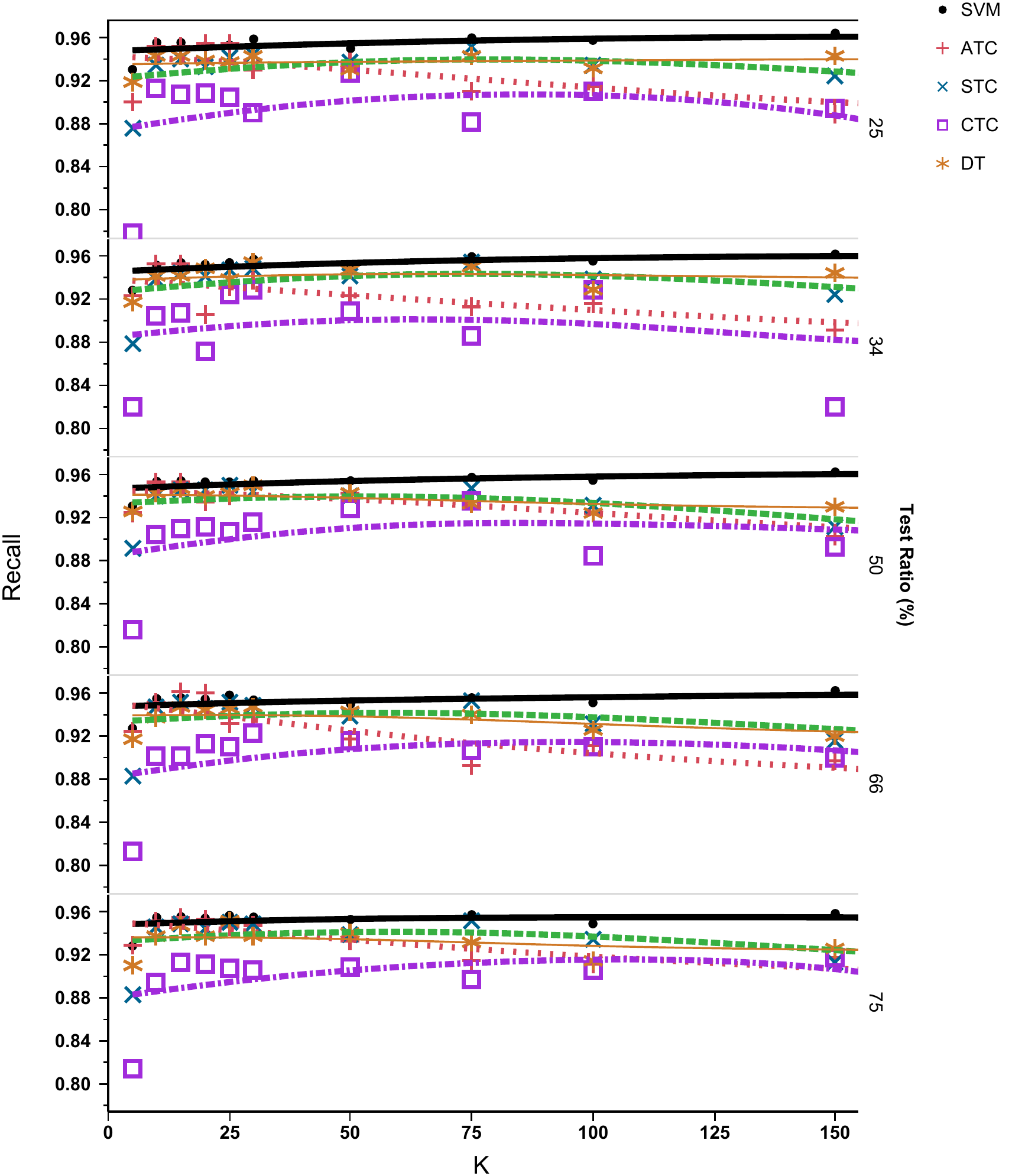}
    }
    \qquad
    \subfloat[F-score
        \label{fig:topicFscoreOrbital}]{
        \includegraphics[width=\smallgraphwidth]{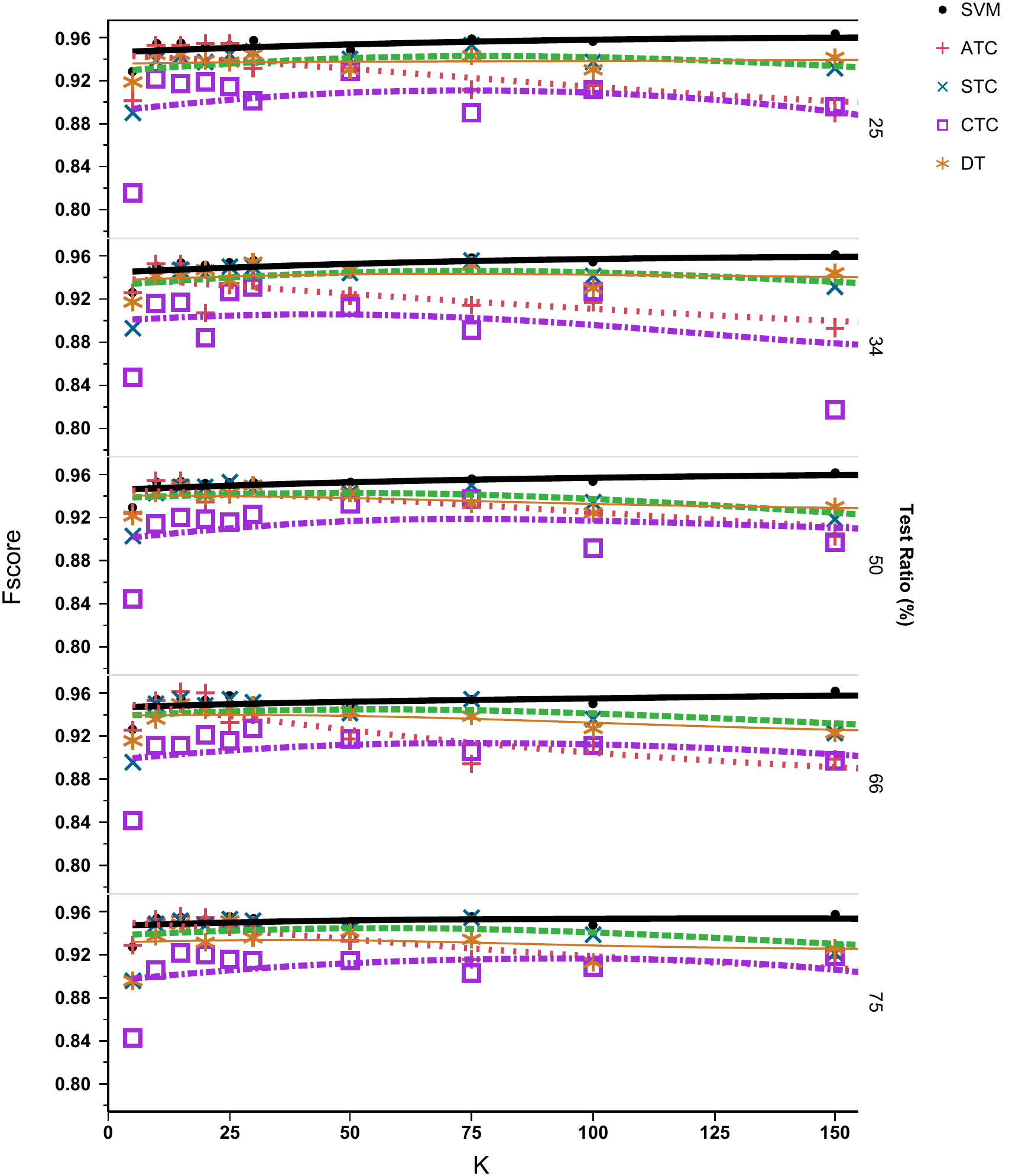}
    }
    \caption{Classification performance using ATC, STC, CTC, DT and SVM for the orbital dataset}
    \label{fig:orbital-all}
\end{figure*}

\begin{figure*}[tp]
    \centering
    \subfloat[Precision
        \label{fig:topicPrecisionPediatric}]{
        \includegraphics[width=\smallgraphwidth]{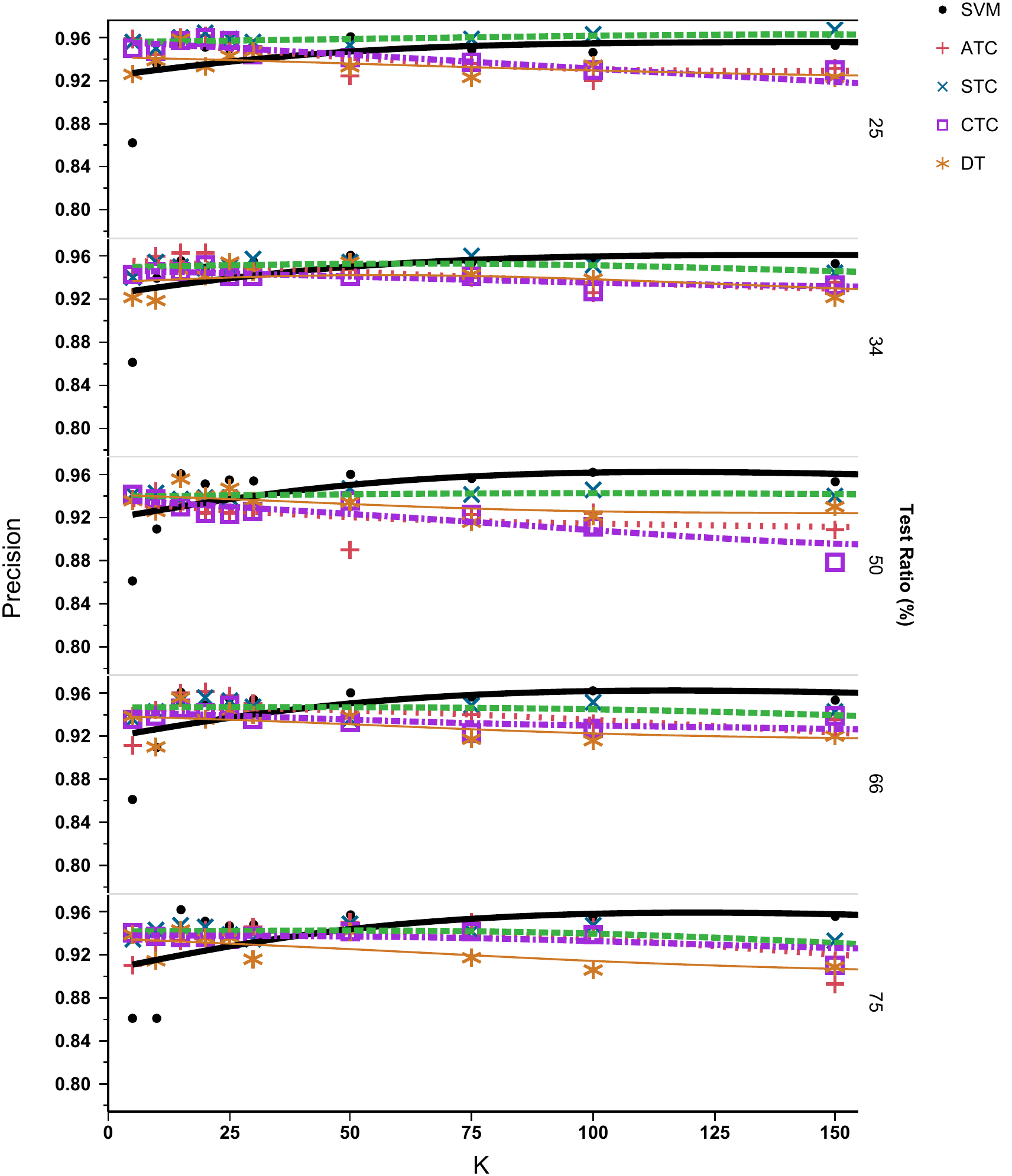}
    }
    \subfloat[Recall
        \label{fig:topicRecallPediatric}]{
        \includegraphics[width=\smallgraphwidth]{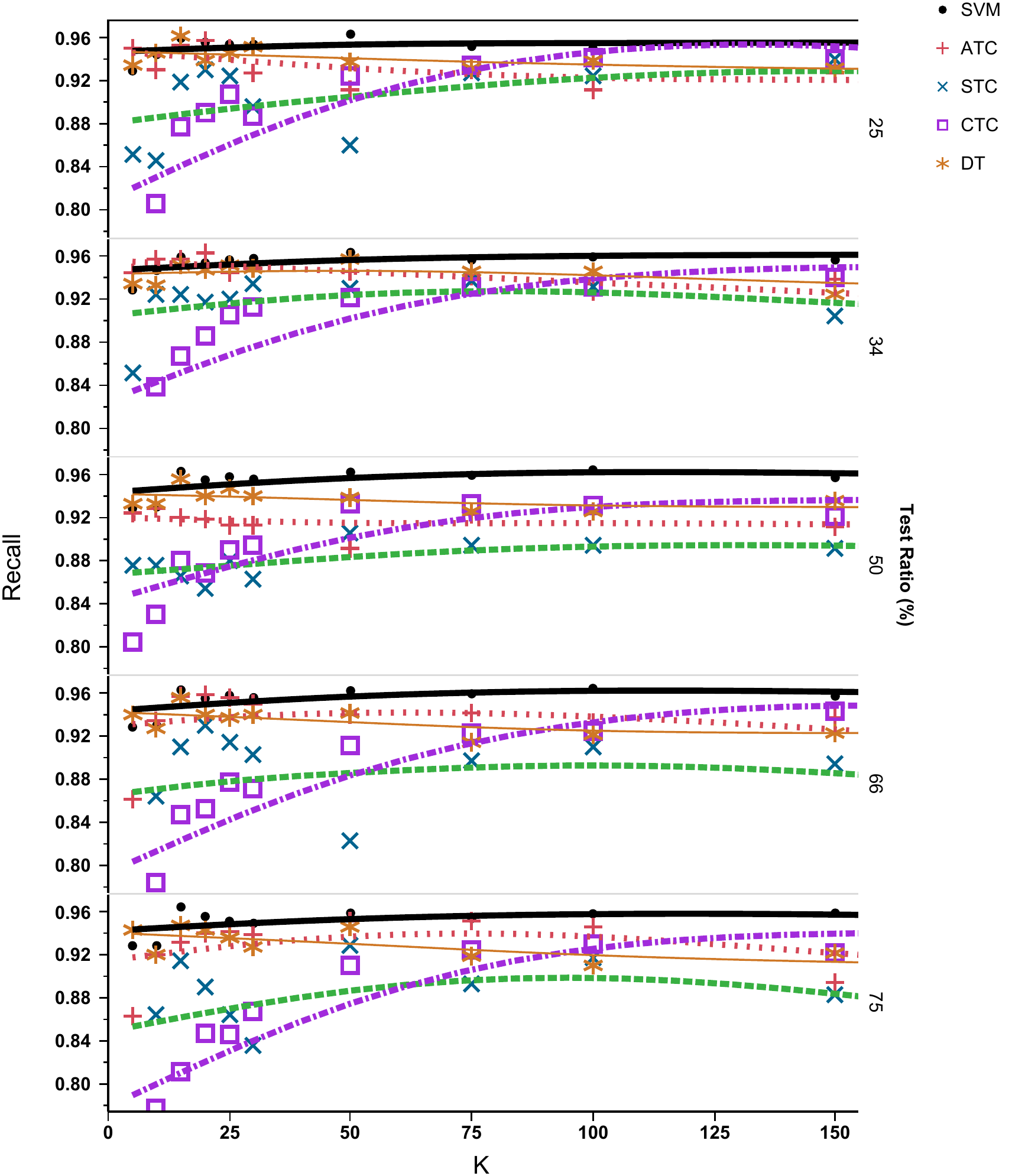}
    }
    \qquad
    \subfloat[F-score
        \label{fig:topicFscorePediatric}]{
        \includegraphics[width=\smallgraphwidth]{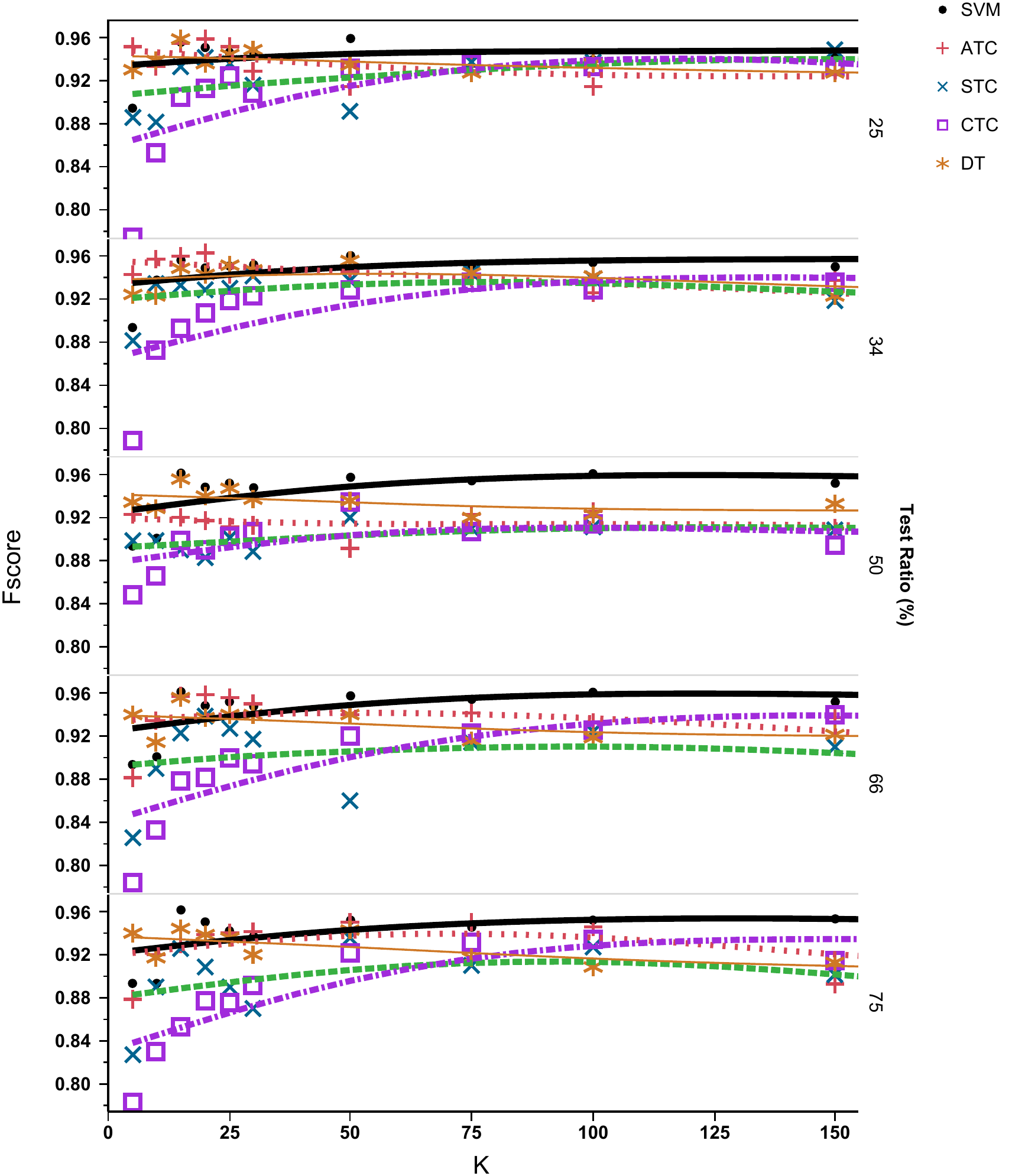}
    }
    \caption{Classification performance using ATC, STC, CTC, DT and SVM for the pediatric dataset}
    \label{fig:pediatric-all}
\end{figure*}

One of the advantages of switching from using the entire vocabulary to represent documents to using topics  as explained in Section \ref{bgr_tm} is the dimension reduction \eqref{eq:dim_red} achieved by this transformation. 
\begin{equation}\label{eq:dim_red}
   Dimension Reduction (\%) = { {\sum attributes - \sum topics} \over \sum attributes }
\end{equation} 
Typically, the vocabulary of a text corpora has a vocabulary in thousands whereas the total number of topics is usually in lower hundreds. The orbital and pediatric datasets had 1,295 and 1,501 attributes respectively. These numbers reflect the total number of attributes after preprocessing such as removal of frequent and infrequent words. For topic numbers ranging from 5 to 150, a dimension reduction of 88\% to 99\% is achieved for the orbital dataset. Similarly, for pediatric dataset, 90\% to 99\% dimension reduction is achieved.
\\
\\
\noindent
Classification performance of ATC, STC and CTC was compared to SVM and decision tree in Figures \ref{fig:orbital-all} and \ref{fig:pediatric-all} for orbital and pediatric datasets respectively. They are each divided into five sections to show the result of using different training/testing proportions. These training and test datasets are randomized and stratified to make sure each subset is a good representation of the original dataset  as explained in Section \ref{sec:exp_dataset}. Also, since the best number of topics is not known in advance, different values were considered ranging from 5 to 150. 
\noindent
Among all techniques, topic vector classification with SVM performed the best especially with higher number of topics. However, for smaller number of topics, ATC and topic vector classification with decision tree performed better or comparable depending on the training dataset size. Having better performance with lower number of topics is desirable as it leads to faster training and testing times. CTC and STC showed varying success depending on the number of topics and training dataset size; CTC showed improvement as number of topics increased since it uses the entire topic vector for classification. ATC and STC, on the other hand, did not improve as much with the increasing number of topics; since they use a single discriminative topic. Finally, different training and testing proportions had little effect on the classifiers' performance for both datasets. This implies that the classifiers generalize well and using only small portion for the training dataset would be sufficient to build an accurate classifier. This is a great outcome as typically, it is difficult to find big labeled datasets as the labeling process is costly.  

\noindent
To summarize, raw text classification using both decision tree and SVM performed well, with SVM performing better than decision tree for both of the datasets. Alternatively, when topic vector representation of the reports were used, the classification performance got better 
for both datasets. Between decision tree and SVM, SVM performed better for topic vector classification as well. Among the topic modeling-based classifiers, ATC performed the best 
for both datasets. 
ATC also performed better than raw text classification but not better than topic vector classification using SVM.
Since ATC is a simpler algorithm compared to SVM, once the topic model is built it may be preferable to use ATC. 
\section{Discussion}
\label{related_work}
\noindent
Other than standard topic modeling techniques, there have been studies to further enhance the capabilities of standard topic modeling. In \cite{wallach06},  Wallach extended the LDA algorithm to handle n-grams. Griffiths et al. combined LDA with POS tagging to have both content and functional words \cite{griffiths2004integrating}.  These studies resulted in a more complex topic-modeling algorithm mostly to make topic-modeling features comparable to Natural Language Processing (NLP), which can slow down the system. Other NLP-based classification techniques, e.g., \cite{Yadav2013a, yadav2016automated},  an be effective in classifying clinical reports as well, however they are computationally expensive and they may require customization by medical experts. As such, we wanted to build a fast and efficient solution using topic modeling without increasing the algorithmic complexity or time to generate topic models. Accordingly, our solutions are based on standard topic modeling algorithms but further extended with our classification techniques. 
\\
\\
In the field of text classification, topic modeling techniques have been used in various ways. Zhang et al \cite{Zhang:2008:EFS:1395080.1395499} used topic modeling as a keyword selection mechanism by selecting the top words from topics based on their entropy. In our study, we removed the most frequent and infrequent words to produce a manageable vocabulary size but we did not use topic model output  as a keyword selection mechanism. Sriurai \cite{Sirurai-2011} compares BoW representation to topic model representation for classification using varying and fixed number of topics respectively. This is similar to our topic vector classification results with SVM. However, because the number of topics typically is not known in advance, we evaluated different numbers of topics, whereas Sriurai \cite{Sirurai-2011} uses a fixed number of topics.  In another similar study, Banerjee \cite{Banerjee:2008:ITC:1390334.1390546} uses topics as additional features to BoW features for the purpose of classification. In our approaches, we used topic vector representation as an alternative to BoW representation and not as additional features. This way, we can achieve greater dimension reduction. 
\\
\\
Other than text classification, topic modeling techniques have also been used in related tasks. Arnold et al. \cite{Arnold:2010fk} shows an information retrieval system where patients can be queried and compared based on their topic distributions. We also used similarity measures to compute the similarity between a report and a class representative topic distribution; however, it is not query-based and it is for classification purposes. 
\section{Conclusion}
\label{conclusion}
In this study, topic modeling of clinical reports was used with different classification techniques and automated clinical outcomes were compared with conventional machine learning techniques. Compared to bag-of-words representation, classification using topic vectors performed comparably with the additional benefit of dimension reduction and interpretability. 
Several supervised classifiers were built based on topic model of the documents in the training dataset. In confidence-based topic classifier (CTC), the topic with biggest confidence for positive class was used to classify reports in the testing dataset. Alternatively, using a similarity-based topic classifier (STC), to classify a document, its topic distribution was compared in similarity to the average topic distributions of each class
Finally, in aggregate topic classifier (ATC), a single discriminative topic was chosen and used to classify the reports in the testing dataset.  Among these topic modeling-based classifiers, ATC demonstrated the best classification performance, however topic vector classification using SVM was the most successful among all classifiers. Since ATC uses fewer topics and less complex than SVM, it may be still be preferable to use ATC for faster performance with comparable accuracy. 
\\
\\
 Results from this study can have significant impacts on the quality and efficiency of healthcare. First of all, the classifiers built in this study can be used to automatically predict the conditions in a clinical report. They can replace the manual review of clinical reports, which can be time consuming and error-prone. In addition, with the increased accuracy and interpretability they provide, clinicians can have more confidence in utilizing such systems in real life settings. Finally, real world datasets such as the ones used in this study could be more challenging than simulated ones. There could be human errors during manual labeling or physicians may disagree. Therefore, it is critical to get good performance on real world datasets so that the systems could be viable to be used in real world settings. Our proposed classifiers provide promising results to be utilized successfully in such settings.

 \bibliographystyle{elsarticle-num} 
 \bibliography{pubs,references_efsun}

\end{document}